%% file: main.tex
\documentclass[10pt,twocolumn,letterpaper]{article}
\usepackage{resources/cvpr}
\usepackage{times}
\usepackage{epsfig}
\usepackage{graphicx}
\usepackage{amsmath}
\usepackage{amssymb}
\usepackage{subfig}
\usepackage{todonotes}
\usepackage{enumitem}
\usepackage{float}
\usepackage{amssymb}
\usepackage[pagebackref=true,breaklinks=true,letterpaper=true,colorlinks,bookmarks=false,urlcolor=blue]{hyperref}

\cvprfinalcopy % *** Uncomment this line for the final submission

\def\cvprPaperID{****} % *** Enter the CVPR Paper ID here

\setlist[itemize]{itemsep=0.1em, topsep=0.1em}

\ifcvprfinal\fi
\begin{document}

% TO BE COMMENTED OUT ONCE ALL TODOS ARE FINISHED ~NK
% \listoftodos
% \setuptodonotes{inline}

%%%%%%%%% TITLE
\title{CNNtention: Can CNNs do better with Attention?} 

\author{
  Nikhil Kapila\\
  {\tt\small \href{mailto:nkapila6@gatech.edu}{nkapila6@gatech.edu}}\\
  \and
  Julian Glattki\\
  {\tt\small \href{mailto:jglattki3@gatech.edu}{jglattki3@gatech.edu}}\\
  \and
  Tejas Rathi\\
  {\tt\small \href{mailto:trathi9@gatech.edu}{trathi9@gatech.edu}}\\
  \and
  \vspace{-2em}
  \\
  \href{https://github.com/AttentionSeekers/CNNtention}{\texttt{Github Repository}}
  }

\maketitle
%\thispagestyle{empty}

\input{1-abstract}
\input{2-intro-bg-moti-objs}
\input{3-datasets}
\input{5-approach}

\input{6-experiments-and-results}

\input{7-conclusion}
\input{10-appendix}
\input{9-work-division}

\clearpage
\twocolumn
{\small
\bibliographystyle{unsrt}
\bibliography{resources/egbib}
}

\end{document}

%% file: 1-abstract.tex
%%%%%%%%% ABSTRACT
\begin{abstract}
Convolutional Neural Networks (CNNs) have been the standard for image classification tasks for a long time, but more recently attention-based mechanisms have gained traction. This project aims to compare traditional CNNs with attention-augmented CNNs across an image classification task. By evaluating and comparing their performance, accuracy and computational efficiency, the project will highlight benefits and trade-off of the localized
feature extraction of traditional CNNs and the global context capture in attention-augmented CNNs. By doing this, we can reveal further insights into their respective strengths and weaknesses, guide the selection of models based on specific application needs and ultimately, enhance understanding of these architectures in the deep learning community.
\end{abstract}

%% file: 2-intro-bg-moti-objs.tex
\section{Introduction}
%{(5 points) How is it done today, and what are the limits of current practice?}
Convolutional Neural Networks (CNNs) have been the state-of-the-art architecture for image classification for many years and revolutionized the field of computer vision \cite{CNNReview} \cite{CNNReview2}.
Through convolution and pooling layers, CNNs are able to extract, preserve and structure spatial information, making them suitable for dealing with images.
However, the architectural biases introduced through these layers force CNNs to rely on local receptive fields, thereby, among other limitations \cite{CNNOtherLimitations}, hindering the networks from capturing long-range dependencies and global context \cite{CnnLimitations}. 
More recently, Vision Transformers (ViTs) have taken over the leaderboards on image classification tasks (see \cite{ImageNetBest} or \cite{CIFAR10Best}), showing the strength of attention-based mechanisms, which excel at capturing global relationships and thereby overcome some of the limitations of CNNs \cite{ViTsVsCnns}. 

\section{Background/Motivation}
It is important to note that, even though Vision Transformers (ViTs) \cite{vision-transformer} are superior to CNNs and have taken over the leaderboards, there are certain limitations in ViTs that deter its use such as requiring significant computational resources and since deep learning often involves trade-offs, it’s crucial to factor this in when planning to deploy a solution especially on edge devices. Furthermore, ViTs require large datasets to learn well which makes it unsuitable for cases where we do not have large amount of data. This makes a strong case that combining the strength of both architectures can lead to a more robust and versatile model.

This idea has already been explored in the past.
In \cite{SqueezeExcitation}, the authors employ Squeeze-And-Excitation blocks, which encode inter-channel relationships and thereby model channel-wise attention.
The authors of \cite{SpatialTransformers} and \cite{UNet} try a different approach by focusing  on spatial features.
Trying to tackle the shortcomings of these methods, the authors of \cite{CoordinateAttention} specifically focus on how attention-mechanisms can be used to model long-range dependencies.
Newer papers, like \cite{CANet} or \cite{EfficientLocalAttention},  focus on the combination of these ideas and refine them further.

%{(5 points) What did you try to do? What problem did you try to solve? Articulate your objectives using absolutely no jargon.} 
%{(5 points) Who cares? If you are successful, what difference will it make? }
\subsection{What are we doing?}
We add 3 different attention mechanisms within a ResNet20 and observe how CNNs behave. %We first start with adding Self-Attention (SelfAtt) and MultiHead attention (MHA) mechanisms across spatial axes motivated from \cite{AttIsAllYouNeed} \cite{selfatt-gans} \cite{AttentionAugmented} and then move our focus to Convolutional Block Attention Modules (CBAM) that work on both spatial and channel axes. \cite{CBAM}

% \subsubsection{Problems with CNNs}
% This hierarchical build-up is the reason why CNNs succeed in vision tasks. However, some limitations of this approach as follows:
% \begin{itemize}
%     \item \textbf{Does not give dynamic treatment}: CNNs give equal importance to all spatial locations in the image. Problem with this approach is that it may lead to problems with complex images wherein CNNs may focus on the background instead of the object at hand.
%     \item \textbf{Focus on local regions}: Convolutional filters build up features that are local, i.e. they do not capture relationships between distant regions of the input but only between the neighboring regions.
% \end{itemize}

\subsection{Novelty}
While the effectiveness of attention has already been demonstrated across various CNN architectures and datasets by the respective authors. Our approach differs from previous works in mainly 2 ways:
\begin{itemize}
\item\textbf{Attention not added after every convolution}: We do not apply attention after every convolution operation as seen in \cite{CBAM} but rather after a sequence of different convolution operations. We do it this way to achieve an efficient trade-off which enables us to evaluate if we can add attentions with reduced computation overhead. In our architecture, we have added attention only 3 times in the full architecture as seen in \autoref{fig:architecture}.
\item\textbf{No information given about positions}: We apply attention directly to convolutional layers without explicitly incorporating positional encodings unlike other approaches.
\end{itemize}

% Through our qualitative analysis, we hope to provide insights into relative impact of using Attention in feature buildup of CNNs through the use of GradCAM \cite{gradcam}. These findings could provide a basis for future research in augmenting hybrid attention models into deep learning models.

% \subsection{Initial Hypothesis}
% We expect the below to happen based on the previous discussion so far.
% \begin{itemize}
%     \item \textbf{CBAM shall perform better than Self/MHA}: Since CBAM utilizes both spatial and channel attentions compared to only spatially in Self/MHA models, we expect CBAM to perform better.
%     \item \textbf{SelfAtt/MHA/CBAM shall beat baseline} We expect attention augments to beat baseline as attention mechanisms shall refine built-up linear features with attention. 
%     \item \textbf{Multihead shall overpowers SelfAttention}: Multi-Head Attention (MHA) enhances self-attention (SA) by employing multiple MLPs (heads) to approximate Key, Queries and Values. We expect that MHA will be able to capture more diverse patterns and aggregation of these patterns should in theory, enhance the models ability to perform and generalize better.
% \end{itemize}

%% file: 3-datasets.tex
\section{Datasets}
%{(5 points) What data did you use? Provide details about your data, specifically choose the most important aspects of your data mentioned \href{https://arxiv.org/abs/1803.09010}{here}. You don’t have to choose all of them, just the most relevant.}
\subsection{CIFAR-10}
We use CIFAR10 dataset \cite{CIFAR10} as one of our datasets. Krizhevskii et al created the dataset as a reliably labeled subset of the 80 million tiny images \cite{80milliontiny} in order to show how semi-supervised learning and pre-training can improve classification models \cite{Krizhevsky2009LearningML}. 
The 80 million tiny images dataset contained only weak labeled images scraped from the web, so a subset of semantically similar labels was corrected and filtered, resulting in a self-contained data set that consists of 60000 corrected, randomly-selected, colored and labeled 32×32 images with 10 classes of 6000 instances each, which show animals or modes of transportation.
The creators already apply a 50k/10k train/test split, and we evaluate our models on the test set.
For tuning, we employ a 45k/5k train/validation split and pick the best performing model.
We reuse the data augmentation techniques introduced in \cite{deepresiduallearningimage}, meaning that for the training set we randomly flip the images horizontally and take random crops of padded (with a padding of 4) versions of the images.
Both the training and testing set are normalized based on values from \cite{Martyn}.

\subsection{MNIST} \label{para:mnist-dataset}
% read: https://edstem.org/us/courses/60909/discussion/5857027?answer=13555798

% Another dataset we use is the well-known MNIST handwritten digits (see \cite{MNIST}). 
% The dataset combines images of handwritten digits by students and government workers, which were originally collected by the National Institute of Standards and Technology.
% For the creation of MNIST, the original images were pre-processed by reducing their size to 28×28, normalizing and anti aliasing.
% The resulting dataset consists of 70000, handwritten, 28×28 grayscale images of digits between 0 and 9 with pixel values representing intensities.
% In its original version, the dataset contains 4 incorrectly labeled images \cite{MNIST_wrongLabels}.
% The dataset authors already split the dataset into 60000 training samples and 10000 testing samples, which we reuse in our experiments. 
We use the MNIST dataset \cite{MNIST}, a standard benchmark which comprises of 70,000 grayscale images of handwritten digits (0-9), each sized 28×28 pixels. The data was originally collected by the National Institute of Standards and Technology and pre-processed to include resizing, normalization, and anti-aliasing. Pixel values represent intensity levels, and the dataset includes four incorrectly labeled examples \cite{MNIST_wrongLabels}. For more details, refer to the dataset source.

%% file: 5-approach.tex
\section{Approach}
For our baseline, we create an implementation of ResNet-20 (20 layers) inspired from \cite{deepresiduallearningimage}.
% By choosing a smaller sized ResNet, we can show how these mechanisms behave in less complex architectures than the ones used in the original papers.

%{(10 points) What did you do exactly? How did you solve the problem? Why did you think it would be successful? Is anything new in your approach? }
% \textbf{Important: Mention any code repositories (with citations) or other sources that you used, and specifically what changes you made to them for your project. }

\subsection{CIFAR-10 Baseline}
\begin{figure}
\centering
\includegraphics[width=0.5\linewidth]{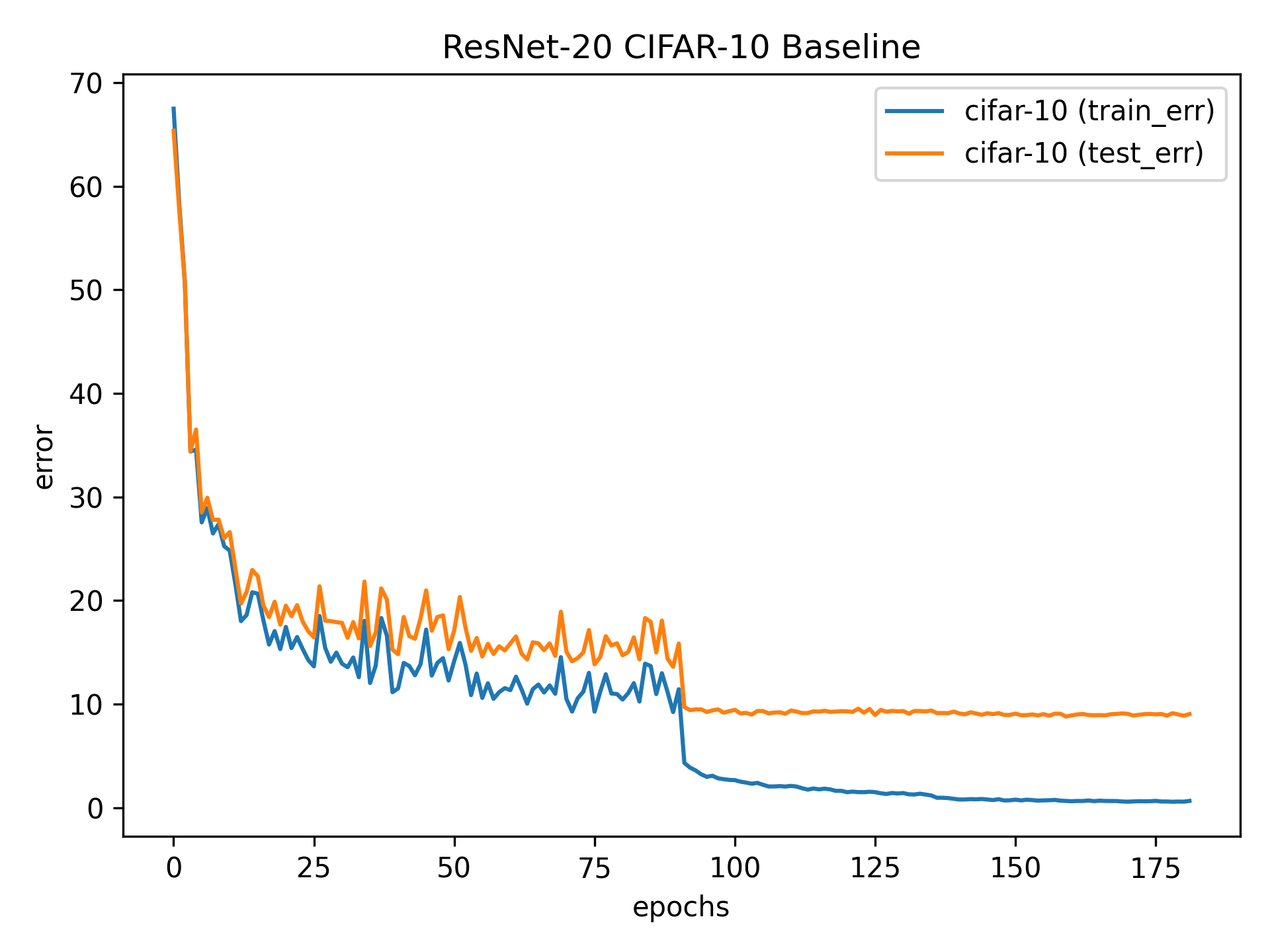}\hfill
\includegraphics[width=0.5\linewidth]{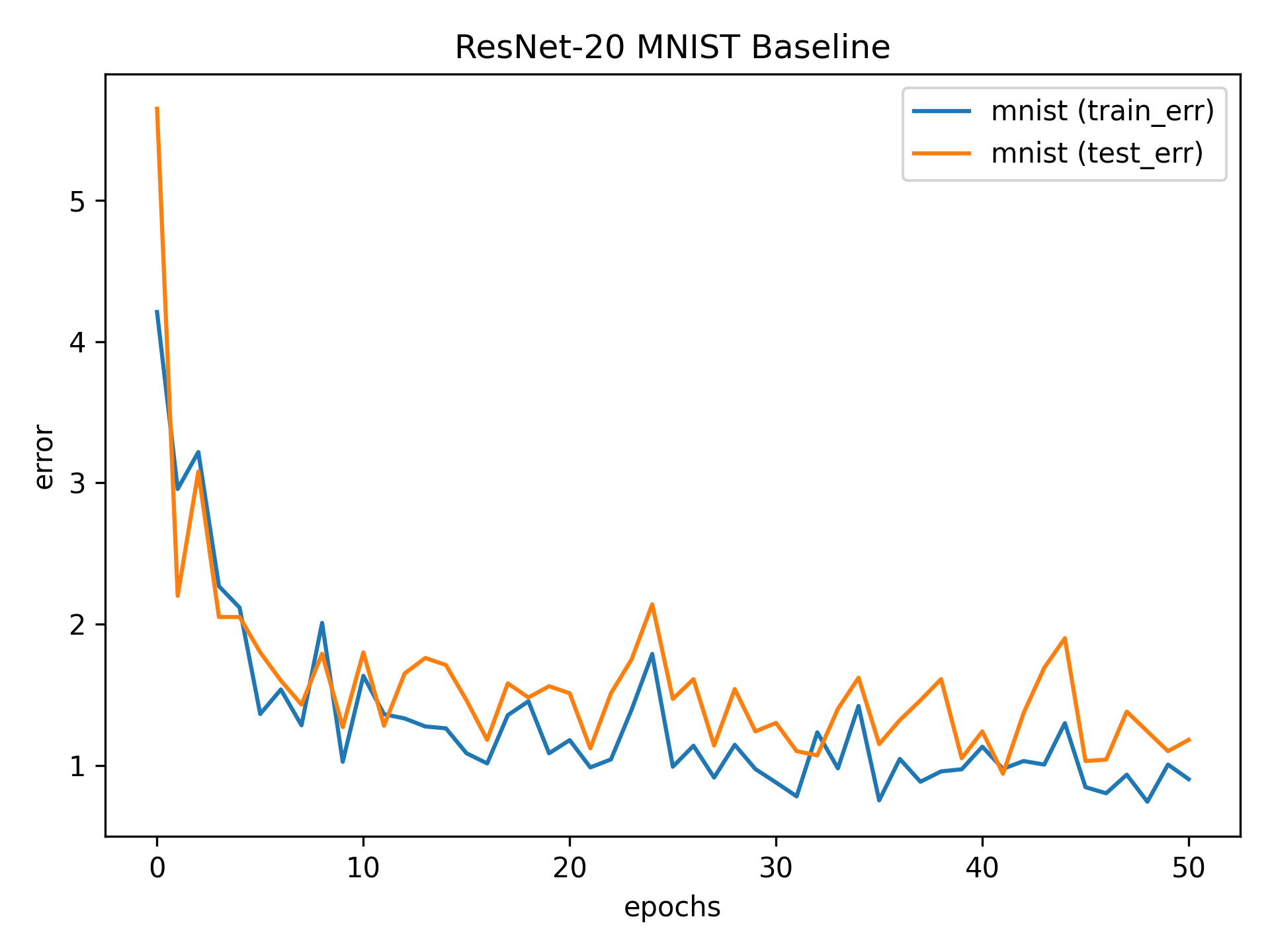}
\caption{CIFAR-10 and MNIST ResNet re-implementation training/testing error.}
\label{fig:resnet-reimplementation-train-test-error}
\end{figure}

First, we reduce the number of output channels in the initial convolution from 64 to 16 and adjust the kernel size, stride, and padding from 7, 2, and 3 to 3, 1, and 1, respectively.
%This is because the authors mention that they use 3×3 convolutions in the first layer and have 16 filters and a feature map size of 32 in the first residual stage.
Second, we remove the subsequent max-pooling operation, as it is not mentioned in the paper.
Third, we lower the number of residual stages from 4 to 3 and use output channels sizes of 16, 32 and 64. %, as this is explicitly mentioned by the authors.
Fourth, in each residual block, when the spatial dimensions and number of channels increase, %we adopt Option A from the paper
which applies identity mapping by padding feature maps with zeros to handle dimension mismatches.
After experimenting with different variations %of Option A, 
we settled on the concise implementation shown in \cite{Idelbayev18a}.
Lastly, we adjusted the final layer to 64 input features, as the final number of output channels is 64, and map these to the 10 classes of CIFAR-10. By training \& evaluating the re implementation with the same hyperparameters mentioned in original paper (see \autoref{tab:original-resnet-hyperparameters}), we achieve a test error of 9.04, closely matching the error of 8.75 mentioned in the original paper. 
Furthermore, as shown in \autoref{fig:resnet-reimplementation-train-test-error}, when training the model on the entire training set as in \cite{deepresiduallearningimage}, we achieve similar training and test error curves compared to the original papers.

\begin{table}[h]
\centering
\begin{tabular}{|l|l|}
\hline
\textbf{Hyperparameter} & \textbf{Value} \\
\hline
Batch Size and Epochs      & 128 batch size, 182 epochs \\
Learning Rate   & 0.1 \\
Schedule (gamma=0.1)      & Milestones: 91, 136 \\
Optimizer and Momentum       & SGD, 0.9 \\
Weight Decay    & 0.0001 \\
\hline
\end{tabular}
\caption{Hyperparameters adapted from \cite{deepresiduallearningimage}. Milestones (91, 136 epochs) correspond to 32k and 48k iterations with a batch size of 128.}
\label{tab:original-resnet-hyperparameters}
\end{table}

\subsection{MNIST Baseline} 
We extend the CIFAR-10 model to MNIST by changing input of initial convolutions from 3-channels to 1-channel and use it as our baseline model. Please note that the only reason we use MNIST is to ease our GradCAM \cite{gradcam} evaluations, baseline results in CIFAR-10 are sometimes confused between the spatial area of interest and background elements. By including a qualitative analysis with the cleaner 1-ch images of MNIST, we aim to provide a clearer evaluation of our experimentation.

%{(10 points) What did you do exactly? How did you solve the problem? Why did you think it would be successful? Is anything new in your approach? }
% \textbf{Important: Mention any code repositories (with citations) or other sources that you used, and specifically what changes you made to them for your project. }

%{(5 points) What problems did you anticipate? What problems did you encounter? Did the very first thing you tried work? }

\subsection{Attention}
In this section, we shall lay the foundation by explaining where attention is being added and how attention is being implemented.

\subsubsection{The feature extractors} \label{sec:feat-exts}
In our baseline implementation, the network is structured into sequential groups of residual blocks referred to as \textbf{layer 1}, \textbf{layer 2}, and \textbf{layer 3} in our code repository \cite{cnntention-repo}.

\begin{figure*}[t!]
\centering
\includegraphics[width=0.8\textwidth]{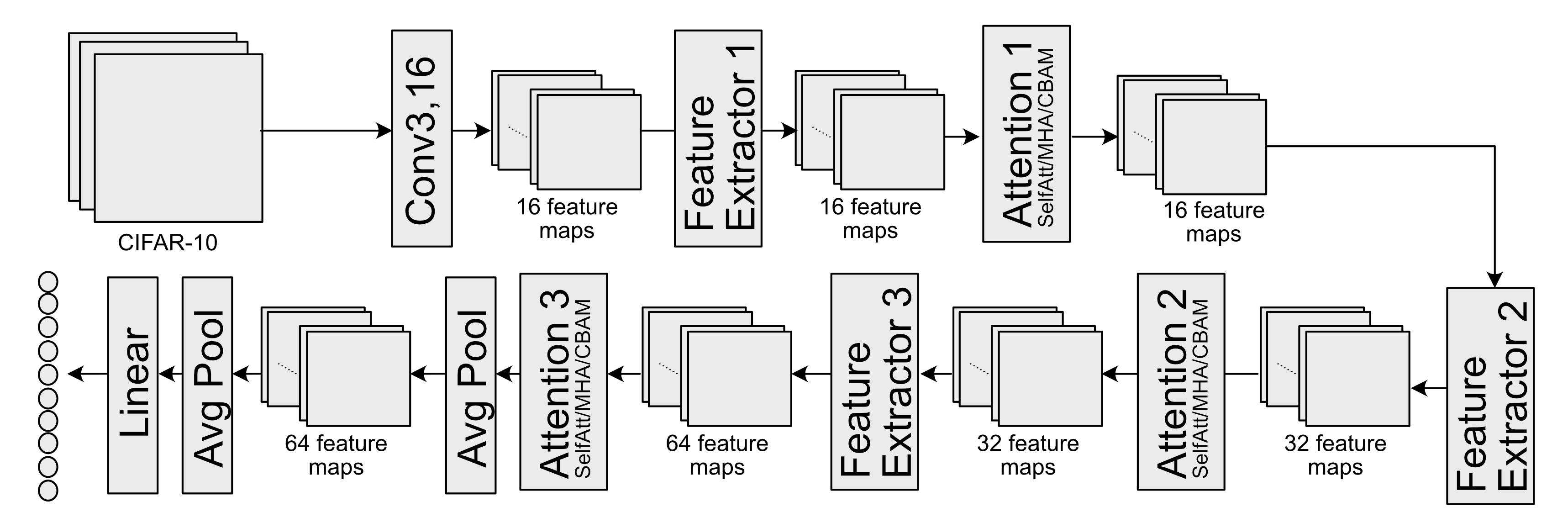}
\caption{Overall architecture with attention added, the Feature Extractor layers are sequential blocks of convolutions that can be seen in \cite{cnntention-repo}. Residual connections are not shown but exist between each feature extractor and attention layer as seen in \autoref{fig:att-augmented}.}
\label{fig:architecture}
\end{figure*}

\textbf{Layer 1} extracts basic features like edges. \textbf{layer 2} reduces spatial resolution, captures mid-level features, and increases filters from 16 to 32 channels. \textbf{Layer 3} reduces spatial resolution and increases feature channels for deeper representation. This can be seen in detail in \autoref{fig:architecture}. We shall further refer to these layers as \textbf{Feature Extractors} 1, 2, and 3. 

\textbf{But where to place Attention? A trade-off:} There is an interesting trade-off that plays out in our choice of adding the attention blocks between the feature extractors instead of within the feature extractors as seen in \autoref{fig:architecture}, i.e. after every convolution layer (within the residual blocks). It is seen in the CBAM paper \cite{CBAM} that adding attention between each convolutional layer could provide more control over feature importance. 

While this may be true, we avoid this for 2 reasons:
\begin{itemize}
\item \textbf{Compute/time/memory}: Applying attention regularly increases computational and memory overhead.
\item \textbf{Noisy features}: Features learned in the early layers may be noisy and not benefit from attention.
\end{itemize}

We let the feature extractors to create meaningful features and then add attention mechanisms to enhance these features. 

\subsubsection{Self-attention (SelfAtt)}
The SelfAtt block enhances spatial relationships (along spatial axes) in feature maps by leveraging self-attention from \cite{AttIsAllYouNeed} but in the context of images.  We start by using 1x1 convolutions to project the input feature maps $F$ into keys (K), queries (Q), and values (V) to create task specific representation of the input features as seen in \cite{AttIsAllYouNeed} \cite{AttentionAugmented}.

\begin{figure}
\centering
\includegraphics[width=\linewidth]{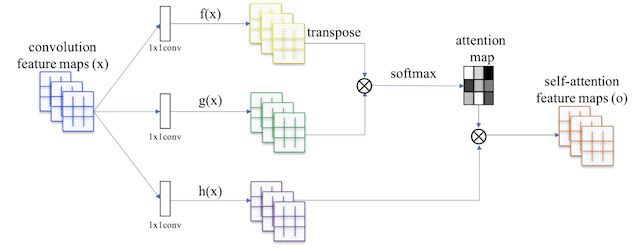}
\caption{Self Attention module introduced in Self-Attention GANs. \cite{selfatt-gans}.}
\label{fig:att-augmented}
\end{figure}

The block then computes raw attention scores (followed by softmax) by computing pair wise similarities which tells us how each spatial position attends to another spatial position. These scores are used to weight the values which combines relevant features across spatial positions.

\begin{align}
\text{SelfAtt}(Q, K, V) &= \text{softmax}\left({QK^\top}\right)V \label{eq:self-att} \\
Q &=XW_Q,\quad K=XW_K,\quad V=XW_V. \notag
\end{align}

\textbf{Using 1x1 convs instead of linear layers:} While the original paper \cite{AttIsAllYouNeed} uses linear layers to model Queries (Q), Keys (K), Values (V), using 1x1 conv firstly helps to preserve the spatial structure of our feature maps $F$ and not lose it by collapsing it for a linear block. 

\textbf{Similarity in information aggregation:} The 1x1 convolution in theory act similar to the global/avg pooling in CBAM (\autoref{sec:cbam}). CBAM's pooling aggregates spatial information globally for channel attention while 1x1 convolutions transform and mix channels locally at each spatial position for self-attention.

\textbf{Omission of $\sqrt{d_k}$:} In \autoref{eq:self-att}, we omit the scaling factor since the input channels (16, 32, 64) are small unlike embeddings in language models that are huge in comparison. Furthermore, omitting the scaling factor aids faster convergence as it would lead to stronger gradients.

\subsubsection{MultiHead (Self) Attention (MHA) Block}
We extend SelfAtt from \cite{selfatt-gans} to MHA motivated from \cite{AttIsAllYouNeed} which uses multiple heads to estimate attention features. We use 8 heads so that MHA can develop a distributed representation in hopes of better approximating long-term dependencies compared to SelfAtt.

\begin{align*}
\text{MultiHead}(Q, K, V) &= \text{Concat}(\text{head}_1, \dots, \text{head}_h)W_O, \\
\text{where } \text{head}_i &= \text{Attention}(QW_Q^i, KW_K^i, VW_V^i), \\
\text{Attention}(Q, K, V) &= \text{softmax}\left(\frac{QK^\top}{\sqrt{d_k}}\right)V, \\
Q &= XW_Q, \quad K = XW_K, \quad V = XW_V.
\end{align*} 

\subsubsection{CBAM Block} \label{sec:cbam}
Convolutional Block Attention Module (CBAM) \cite{CBAM} emphasizes meaningful features along two dimensions: channel and spatial axes. This module supports efficient flow of information within network by identifying features which should be suppressed and emphasized respectively. For the experiments here, channel attention and spatial attention are applied sequentially on a feature map as per below equations. 
\begin{gather*}
    F' = F_c = M_c \times F\\
    F" = F_s = M_s \times F_c\\
    M_c(F) = \sigma ((MLP(F_{avg}^c)) + MLP(F_{max}^c))\\
    M_s(F) = \sigma (f^{7\times7}(F_{avg}^s; F_{max}^s))
\end{gather*}

\begin{figure}
\centering
\includegraphics[width=\linewidth]{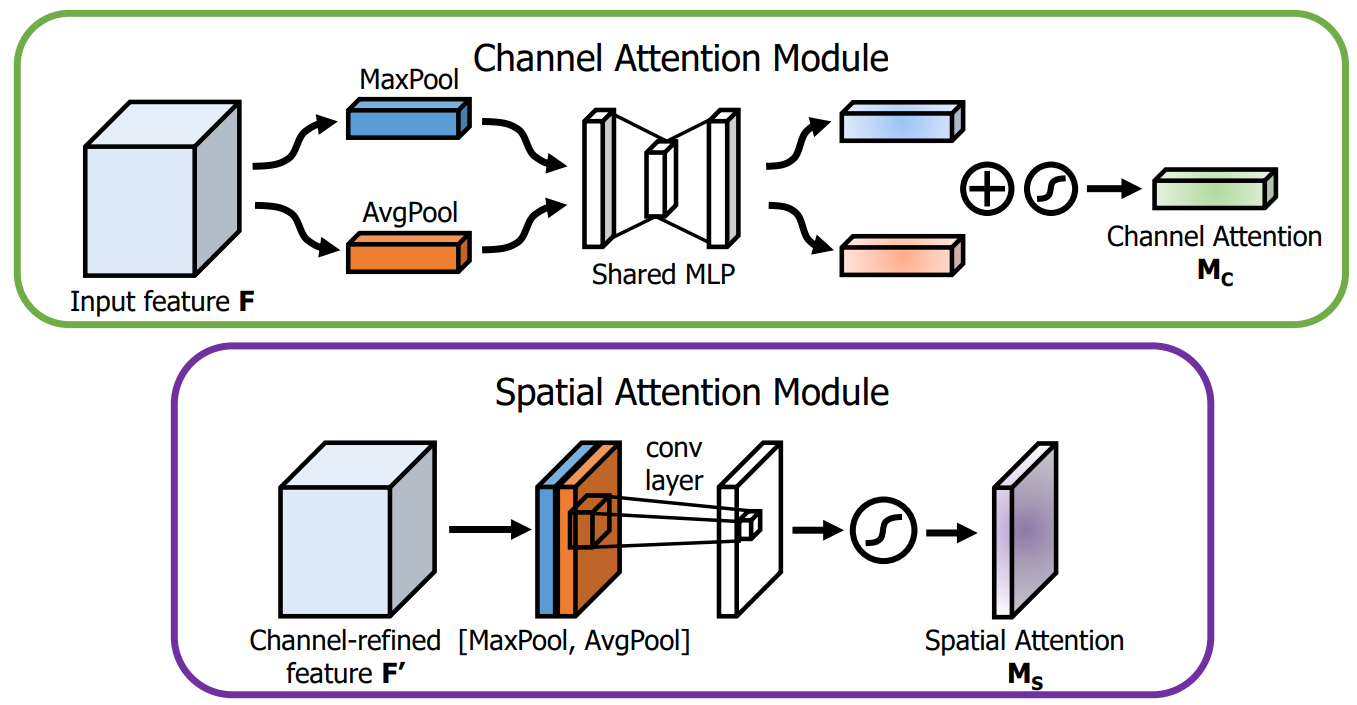}
\caption{CBAM module introduced in \cite{CBAM}.}
\label{fig:cbam-att-module}
\end{figure}

$F \in C \times H \times W$ is the feature map received from feature extractors as seen in \autoref{fig:architecture}.
$F_{avg}, F_{max}$ represents the global average and max pooled features. $F_{avg}^c, F_{max}^c \in {\rm I\!R}^{C \times 1 \times 1}$ are descriptors for channel attention, identified for all channel within feature map of size $H \times W$. This descriptors are forwarded to shared MLP to compute channel attention map $M_c \in {\rm I\!R}^{C \times 1 \times 1}$. $F_{avg}^s, F_{max}^s \in {\rm I\!R}^{1 \times H \times W} $ are the descriptors for spatial attention, identified across channels on feature map of size $H \times W$ which are concatenated and convolved through convolutional layer $f^{7\times7}$ of kernel size 7 to produce spatial attention map $M_s\in {\rm I\!R}^{1 \times H \times W}$. To reduce the number of parameters, number of perceptrons in shared MLP is controlled by a reduction ratio $r$, to produce hidden activation of size ${\rm I\!R}^{C/r \times 1 \times 1}$. The hidden activation is followed by ReLU activation function \cite{CBAM}.

\paragraph{CBAM block placement:} Our implementation differs from the original paper \cite{CBAM} by adding the CBAM block after every residual block instead of within the residual block.

%% file: 6-experiments-and-results.tex
\section{Experiments and Results}

% \textbf{Important: This section should be rigorous and thorough. Present detailed information about decision you made, why you made them, and any evidence/experimentation to back them up. This is especially true if you leveraged existing architectures, pre-trained models, and code (i.e. do not just show results of fine-tuning a pre-trained model without any analysis, claims/evidence, and conclusions, as that tends to not make a strong project). }

All our experimentation are recorded on MLFlow \cite{mlflow}, an open source utility to track experiment runs. Please note that experimentation is done solely on CIFAR-10, we extend the best hyperparameters from CIFAR-10 to MNIST only for GradCAM observations as explained in \autoref{para:mnist-dataset}.

\subsection{Experiments}
\textbf{Optimizer changes, learning rate, regularization:} While SGD uses a more straightforward optimization path, Adam tends to converge faster and is more robust to hyperparameter choices due to its adaptive learning rates based on first and second moment of gradients. Apart from the optimizer, different experiments were ran for each model. We have mostly noted that model has a stable optimization path when learning rates are lower. We further play with regularization to achieve the best generalizable performance and employ EarlyStopping to save on computation costs. This can be seen in detail in our MLFlow results in our repo \cite{cnntention-repo}.

\begin{figure}
\includegraphics[width=0.5\linewidth]{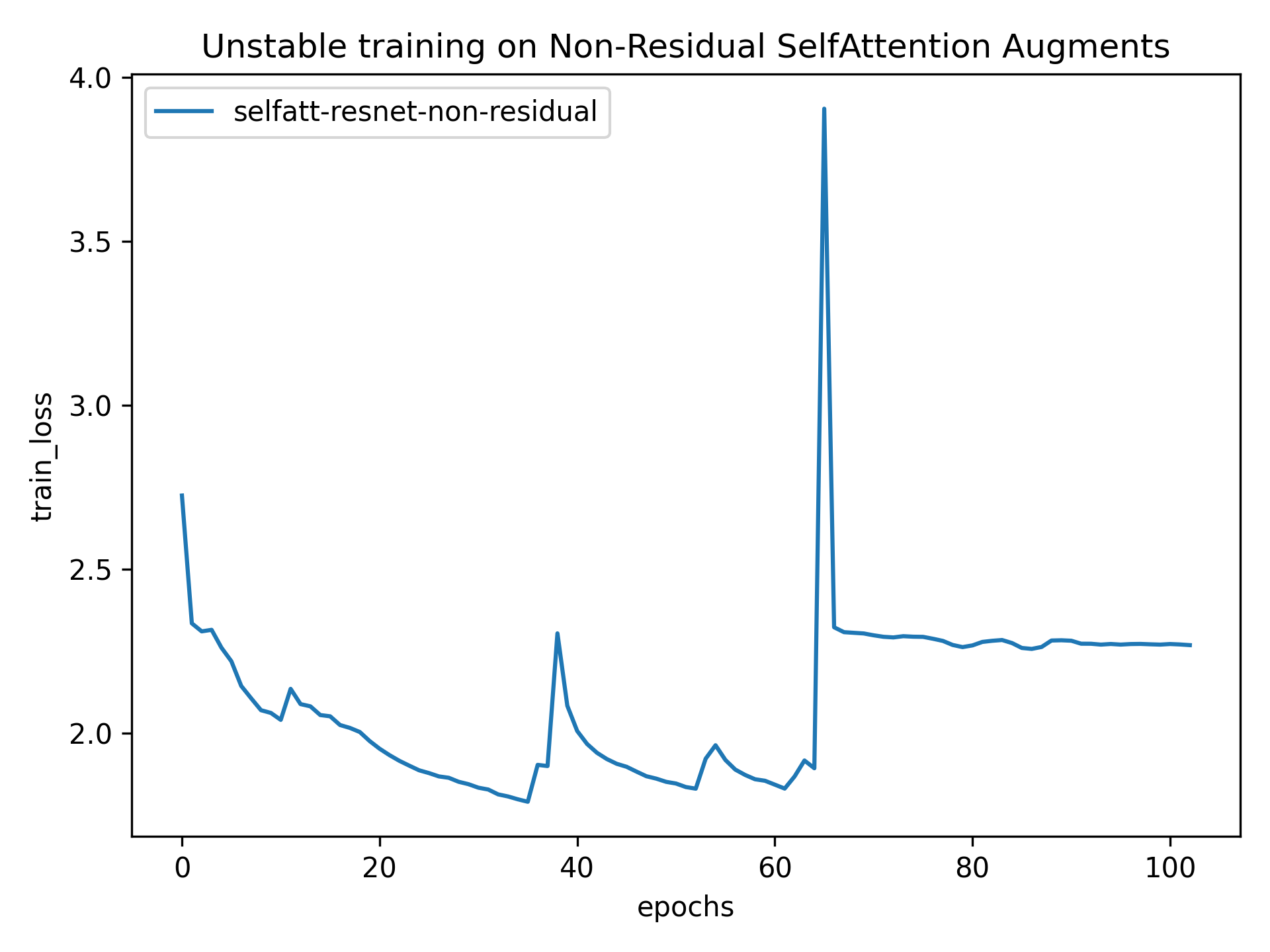}\hfill
\includegraphics[width=0.5\linewidth]{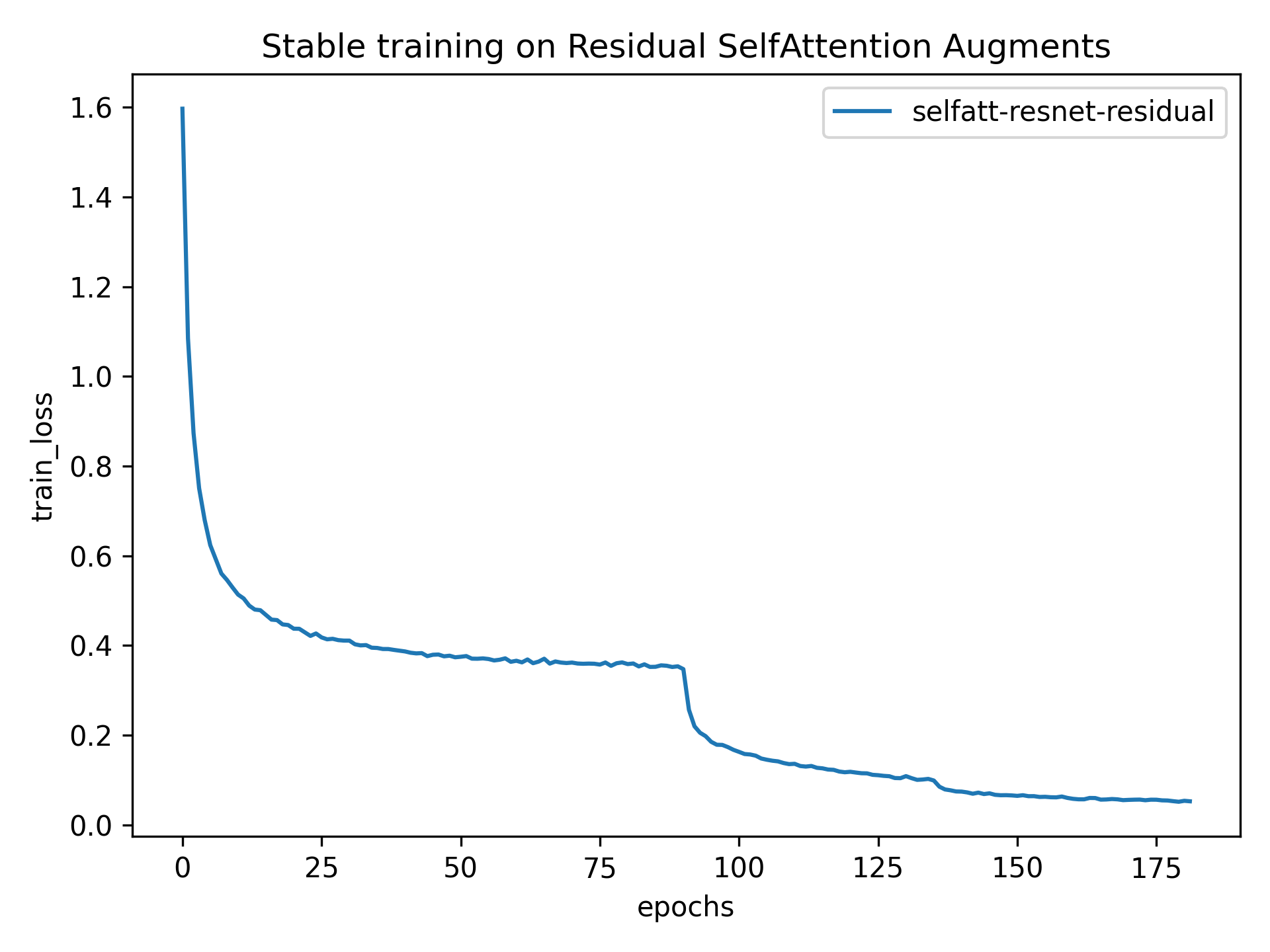}
\caption{Stable training with (right) and without (left)  residual connections between self-attention blocks.}
\label{fig:res-vs-non-res}
\end{figure}

\textbf{The need for residual connections:} \label{para:res-conns} We initially train our models by using SelfAtt blocks without residual connections between the feature extractor layers, exactly as per \autoref{fig:architecture}. However, this led to unstable training as seen in \autoref{fig:res-vs-non-res}. While attention mechanisms are powerful to capture long range dependencies, their contributions during the initial training stages may be limited as they are trying to build up effective representations of Q, K, and Vs.
\begin{align}
R(x) &= F(x) + x, \label{eq:res-connection} \\
\text{where } F(x) &= \text{SelfAtt/ MHA/ CBAM for \autoref{eq:res-connection}}\notag \\
W(x) &= w*F(x) + x, \label{eq:weighted-att} \\
\text{where } F(x) &= \text{SelfAtt/ MHA for \autoref{eq:weighted-att}}\notag
\end{align}

\begin{figure}[h]
\centering
\includegraphics[width=\linewidth]{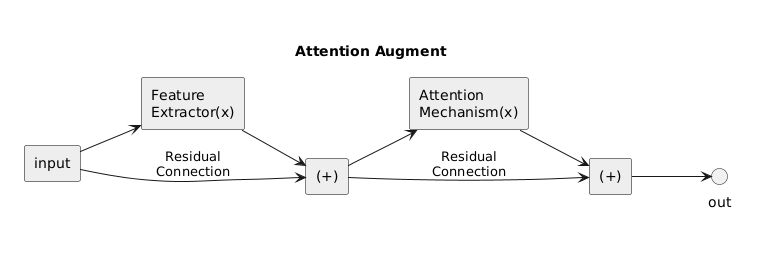}
\includegraphics[width=\linewidth]{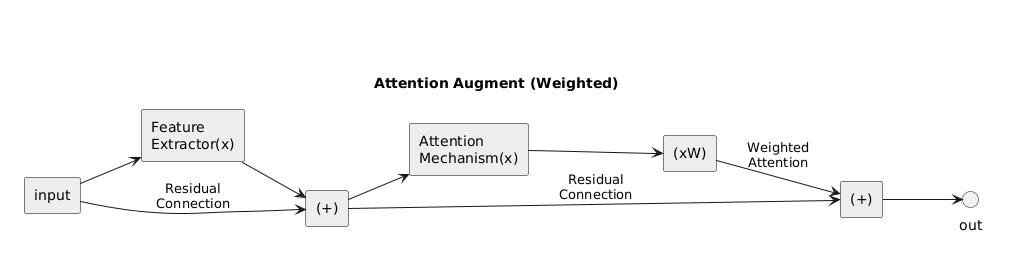}
\caption{Attention added between feature extractor layers utilizing residual connections (top) and weighted residual connections (bottom).}
\label{fig:att-augmented}
\end{figure}

To address this, we introduce residual connections alongside the attention additions as seen in \autoref{eq:res-connection}, \autoref{eq:weighted-att}, and \autoref{fig:att-augmented} which would help propagate information across the feature extraction layers and bypass the attention blocks when necessary. Over time as attentions begins to contribute meaningful representations, the residual connections would act as a stabilizing mechanism ensuring both direct and context information are leveraged efficiently.

This approach was inspired from \cite{AttIsAllYouNeed} and is seen again in Vision Transformers (ViTs) from \cite{vision-transformer} wherein residual connections play a crucial role to stabilize training. While we see this aspect in \autoref{fig:res-vs-non-res} for SelfAtt, we extend the same to MHA and CBAM.

\textbf{Model-Driven feature selection} Inspired from weighted residual concepts for deep networks \cite{weighted-residual}, we train 2 models: with and without weighted attention as seen in \autoref{eq:weighted-att} and \autoref{fig:att-augmented}.

\begin{figure}
\includegraphics[width=0.5\linewidth]{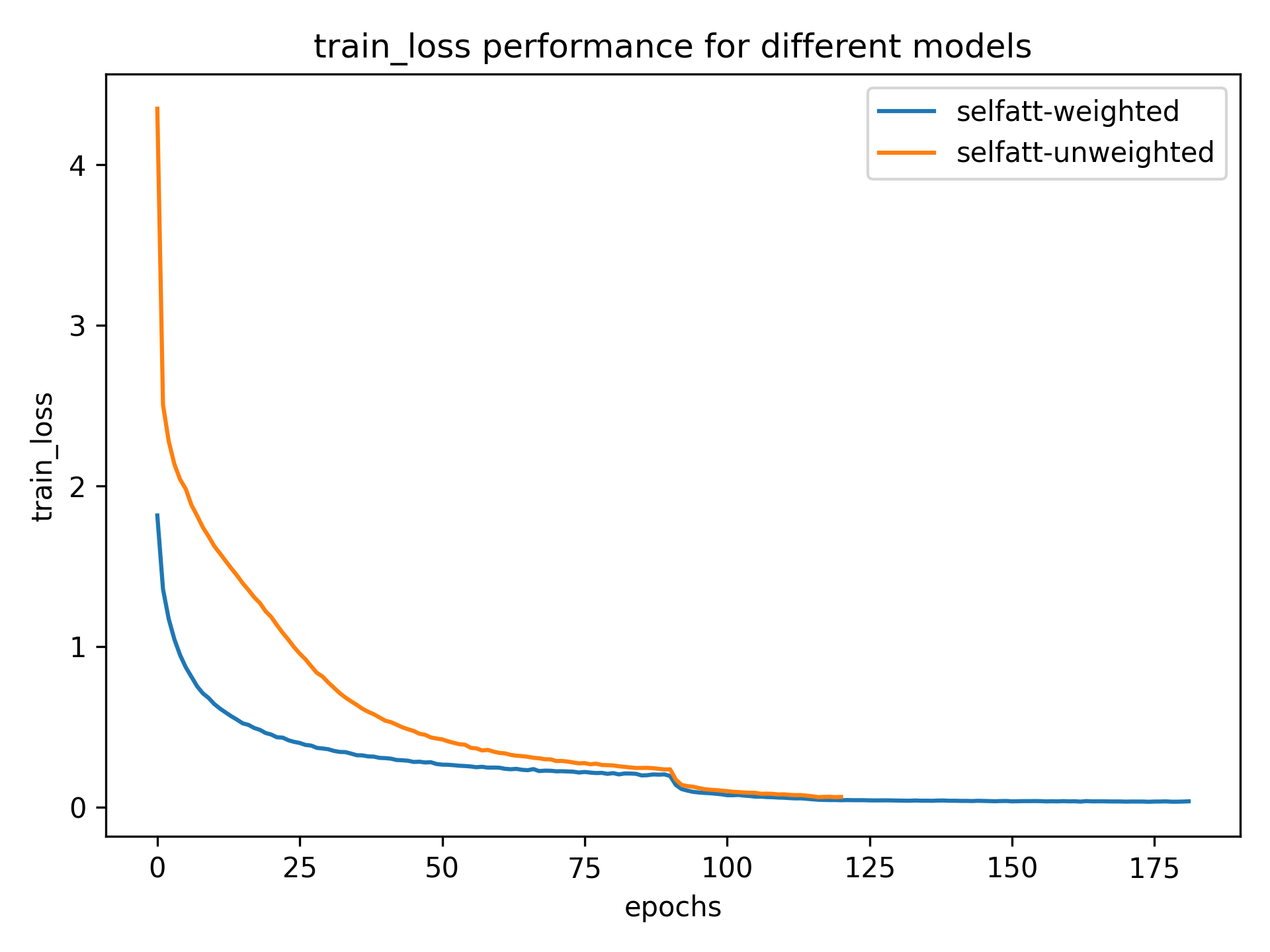}\hfill
\includegraphics[width=0.5\linewidth]{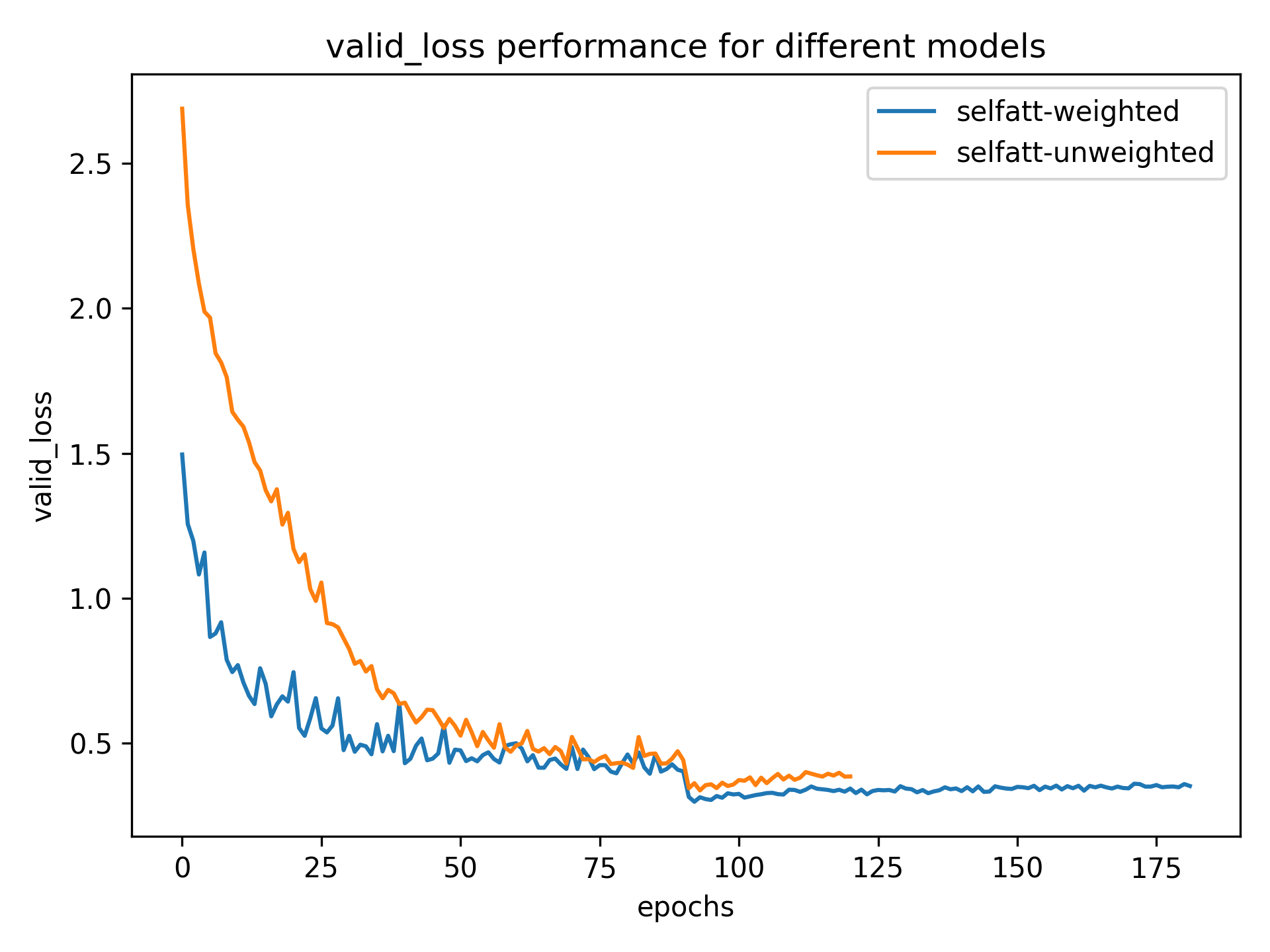}
\caption{Faster convergence when SelfAtt model is allowed to select features dynamically.}
\label{fig:model-driven-feat-selection}
\end{figure}

The model converges faster when it is allowed to dynamically choose between output feature maps (from feat. extractors) and attention based features. During training we see the weighted version of the model has a higher validation accuracy of 90.01\% as compared to 88.00\%.

%When the model is allowed to dynamically choose between original feature extractor outputs or attention based features, it converges faster.
% Moreover, during training we observe that weighted version resulted in validation accuracy of 90.01\% as compared to 88\% for unweighted version. It presents itself as a balance between direct feature propagation (via the residual connection) and the attention-driven transformations.

These weights being zero initially during training directly propagates the output feature maps. This is in line with our discussion from the residual connection section, i.e. self-attention takes time to build up meaningful representations \cite{vision-transformer} \cite{AttIsAllYouNeed}. This is further evidenced in our plots where loss is lower for weighted model as seen in \autoref{fig:model-driven-feat-selection}.

\textbf{SelfAtt hyperparameters extended to MHA:} 
We use the tuned hyperparameters from SelfAtt for MHA as each head in MHA plays a role analogous to SelfAtt. Distribution of representation building across heads should reduce the optimization burden and ensures our comparison is solely due to architectural advantages in MHA.

% We extend the tuned hyperparameters from SelfAtt to MHA since each head in MHA performs a role which is analogous to SelfAtt. Since the load of building up representations is now distributed across multiple heads, we expect that the optimization challenge for each head is simplified compared to a single head model that handles all the tasks alone. Plus, using the same hyperparameters ensures that the comparison (improvement or not) is due to the architectural advantages of MHA. To be explicit, we also carry forward our understanding from the above work of weighted attention and model-driven feature selection in MHA.

\textbf{CBAM's reduction ratio:}
The reduction ratio controls the intermediate dimension of the shared MLP used in channel attention. A smaller reduction ratio has a large hidden layer, more parameters and in turn, higher capacity whereas a large reduction ratio has fewer parameters and lower capacity. The reduction ratio $r$ not only controls the parameter overhead but also how the channel attention block decides to propagate the output. This effect acts as a bottleneck, i.e. a higher reduction $r$ leads to lower number of channels $C/r$ which forces the MLP to retain the most useful information.
In our experiments as seen in \autoref{tab:model-performance-cbam-reduction-ratio}, we tune 3 models with different reduction ratios. We see that model performs similarly across different $r$ with minuscule differences in performance. It means that the shared MLP is able to extract relevant information from the feature maps $F$ even at lower capacities ($r$=32). We strike a balance and pick a value of $r=16$.

\begin{table}
\centering
\begin{tabular}{|c|c|c|}
\hline
\textbf{CBAM-8} & \textbf{CBAM-16} & \textbf{CBAM-32}\\
\hline
90.66\% & 90.32\% & 89.94\% \\
\hline
\end{tabular}
\caption{Comparison of performance on different $r$ of CBAM's attention block.}
\label{tab:model-performance-cbam-reduction-ratio}
\end{table}

%{(10 points) How did you measure success? What experiments were used? What were the results, both quantitative and qualitative? Did you succeed? Did you fail? Why? Justify your reasons with arguments supported by evidence and data.}

\subsection{Results}
We base our success on 2 criterion: (1) better or comparable performance to the baseline \& (2) can our attention added models capture global dependencies? 

\subsubsection{Analysis of results}
\begin{table}
\centering
\begin{tabular}{|l|c|c|c|}
\hline
\textbf{Model} & \textbf{CIFAR-10} & \textbf{MNIST} &\textbf{dur} \\
\hline
Baseline  & 90.96\% & 98.52\% & 3.7h\\
SelfAtt  & 91.44\% & 99.30\% & 2.9h \\
MHA       & 91.06\% & 98.84\% & 5.1h \\
CBAM-16      & 91.32\% & 98.00\% & 1.8h\\
\hline
\end{tabular}
\caption{Comparison of model performance on CIFAR-10 and MNIST datasets. Duration is on CIFAR-10 dataset only, MNIST has similar trends.}
\label{tab:model-performance}
\end{table}

SelfAtt/MHA consistently outperform the baseline showing effectiveness of attention mechanisms on both datasets, i.e. SelfAtt/MHA maintain a better balance between local and global information compared to CBAM. CBAM's final performance trails behind the other attention mechanisms and stabilizes to a higher test error compared to indicating less generalizable performance.

\textbf{Why CBAM trails against SelfAtt?} CBAM is highly focused on suppressing irrelevant feature maps through it's shared MLP \ref{fig:cbam-att-module} and enhancing the critical areas, so this may cause over-suppression of areas of interest. Since CBAM combines 2 attention mechanisms sequentially (channel and spatial), this could also lead to bottleneck effects. To conclude, CBAM may excel at refining feature maps and suppressing irrelevant details but it can lead to loss of global context.  

\begin{figure}
\includegraphics[width=0.5\linewidth]{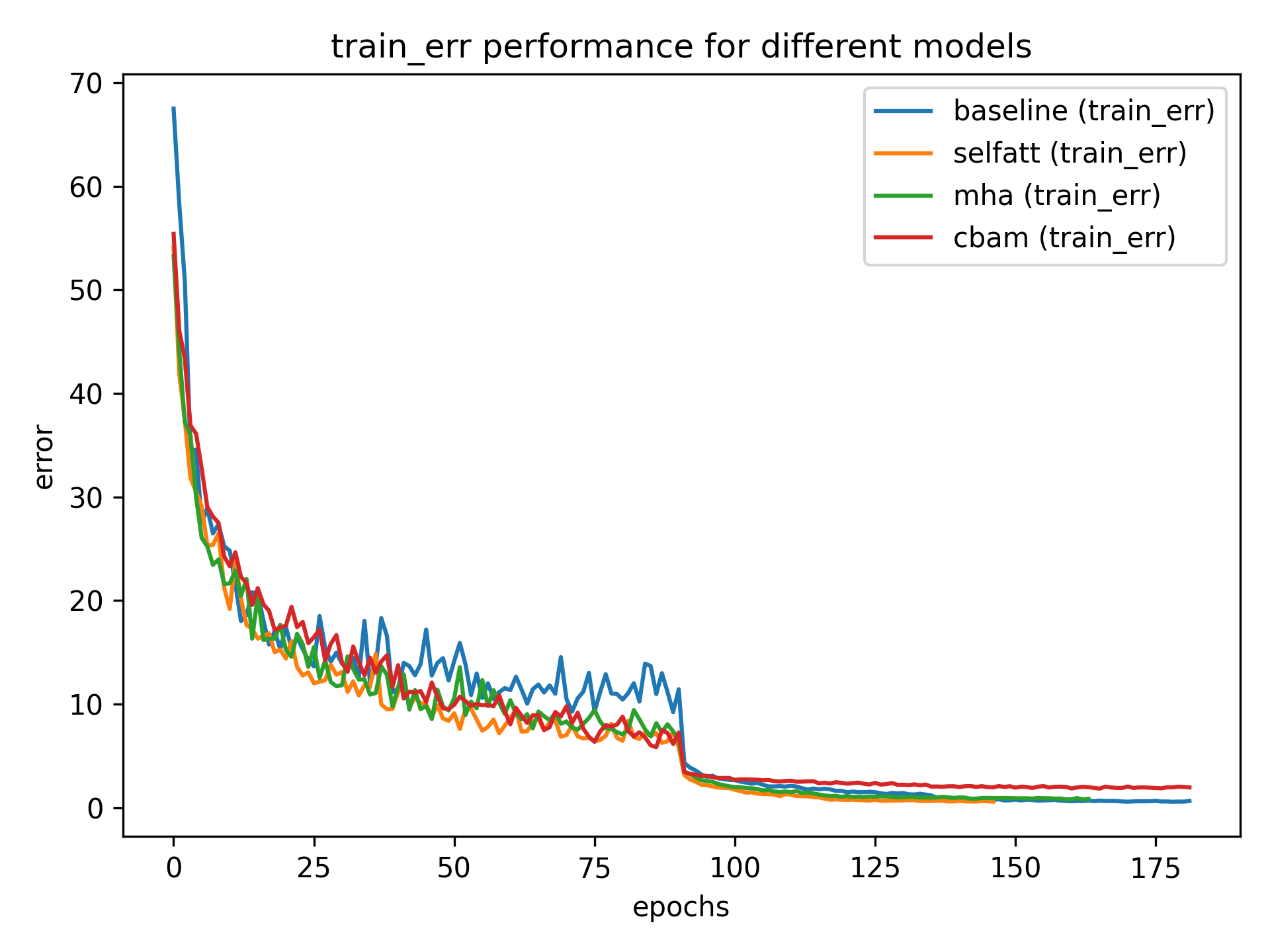}\hfill
\includegraphics[width=0.5\linewidth]{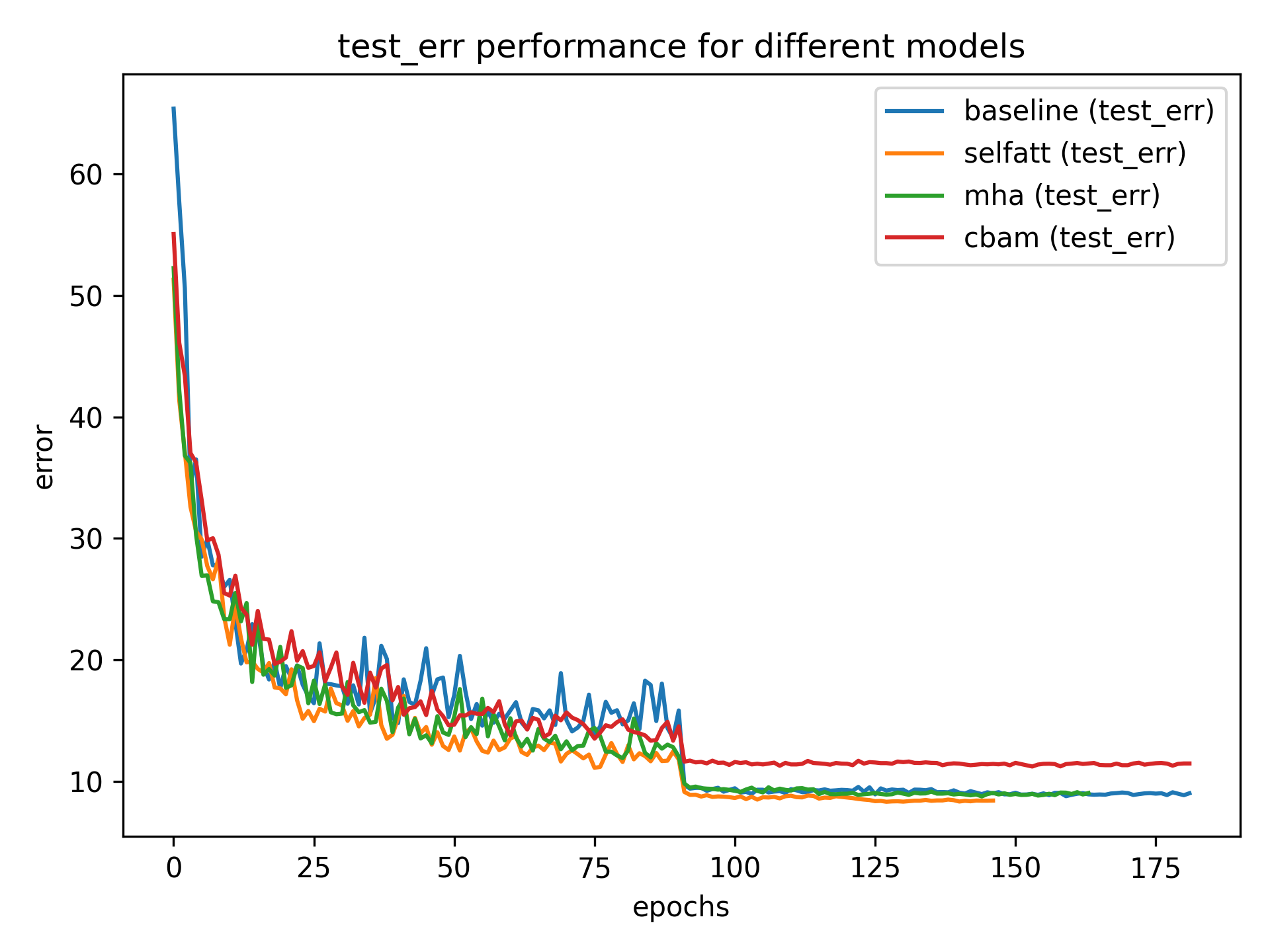}
\caption{Train and test error on different models on CIFAR-10.}
\label{fig:final-comparisons}
\end{figure}

\textbf{CBAM's efficiency trade-off:} It is important to note the trade-off here. While CBAM has comparable performance and in some cases is even better against baseline/SelfAtt/MHA as seen in \autoref{tab:model-performance}, it is the fastest to converge. It shows faster convergence during the initial epochs as seen in \autoref{fig:final-comparisons}. Compared to SelfAtt/MHA, CBAM is lightweight, has lesser parameters to train (only a shared MLP and convolution layer), and hence, has lower computational overhead. To quantify this aspect, we use A-100s to train our MHA in 5 hours, training the same MHA on T4 would require 7.5-8 hours.

\begin{figure}
\centering
\includegraphics[width=\linewidth]{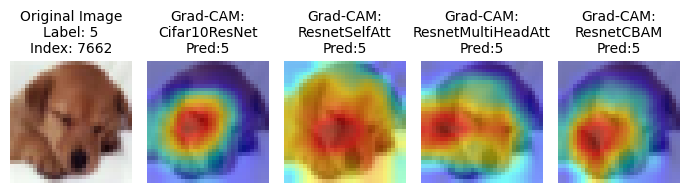}
\includegraphics[width=\linewidth]{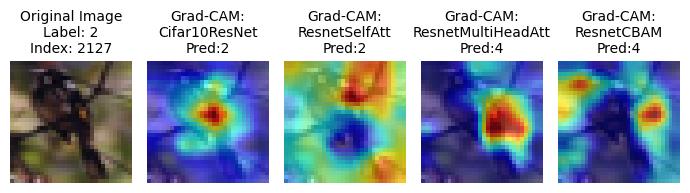}
\includegraphics[width=\linewidth]{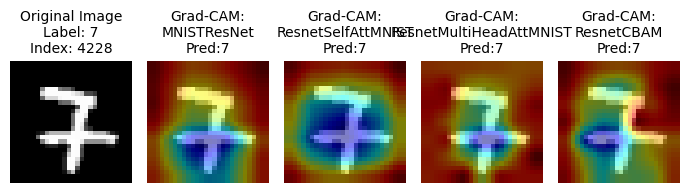}
\caption{GradCAM results on CIFAR-10 and MNIST.}
\label{fig:gradcams}
\end{figure}

\textbf{GradCAM discussion:} Referring to dog image (first row) in \autoref{fig:gradcams}, we can see that SelfAtt effectively models global dependencies. The distributed representation approach in MHA tries to do as good as SelfAtt but it may be possible that each head in MHA learns different representations of Q, K, V which may not be as effective as a single head approach.

The CBAM's channel block first finds the weighted output of the channels. These are used by the spatial block to highlight which parts of the image are important. Hence, CBAM identifies the important regions globally whereas SelfAtt/MHA identifies relationships between all positions. We can see this evidenced in \autoref{fig:gradcams} where CBAM refines the baseline with sharper and effective features but SelfAtt/MHA does so with more granularity by capturing fine grained dependencies between positions (through pairwise computations).

As discussed in \autoref{sec:feat-exts}, we theorized our models to only be able to refine what the baseline model is able to extract. The baseline can extract useful features when there is a contrast between the area of interest and background. There are examples when this is not the case as seen in the bird image (second row). To reinforce this idea, we take the MNIST dataset. In this case, each of models capture global dependencies more effectively as the baseline model can extract better features. More such examples can be seen in \autoref{fig:gradcam-cifar-appen} and \autoref{fig:gradcam-mnist-appen}.

%% file: 7-conclusion.tex
\subsection{Conclusion}
We can conclude that CNNs indeed learn better when attention is added. This is proved by our quantitative and qualitative results. However, each attention mechanism has its own tradeoff to consider in regards to how they model global dependencies and the added computational overhead.

\subsection{Related and Future Work}
This work can be extended in many different directions.

\begin{itemize}
\item\textbf{Further study:} More study can be done to look at attention matrices in SelfAtt/MHA and what weights learn at different reduction ratios for CBAM. Furthermore, SelfAtt/MHA could be applied on both channel and spatial axes to make a 1:1 comparison with CBAM.
\item\textbf{Parallel attention paths:} Inspired from GoogLeNet's inception module \cite{googlelenet}, different attentions can be combined and concatenated to better approximate global dependencies.
\item\textbf{Train attention only:} One could freeze the baseline and only train attention to compare with our results.
\end{itemize}

%% file: 10-appendix.tex
\clearpage
\onecolumn
\section{Appendix}
Here are more GradCAM results. We can see that the SelfAtt model captures long-range dependencies well. MHA's distributed head representation takes the baseline results and generates a more focused global context compared to the baseline. CBAM tends to enhance existing baseline closely with more sharper focus.

\begin{figure}[h]
\centering
\includegraphics[width=0.8\textwidth]{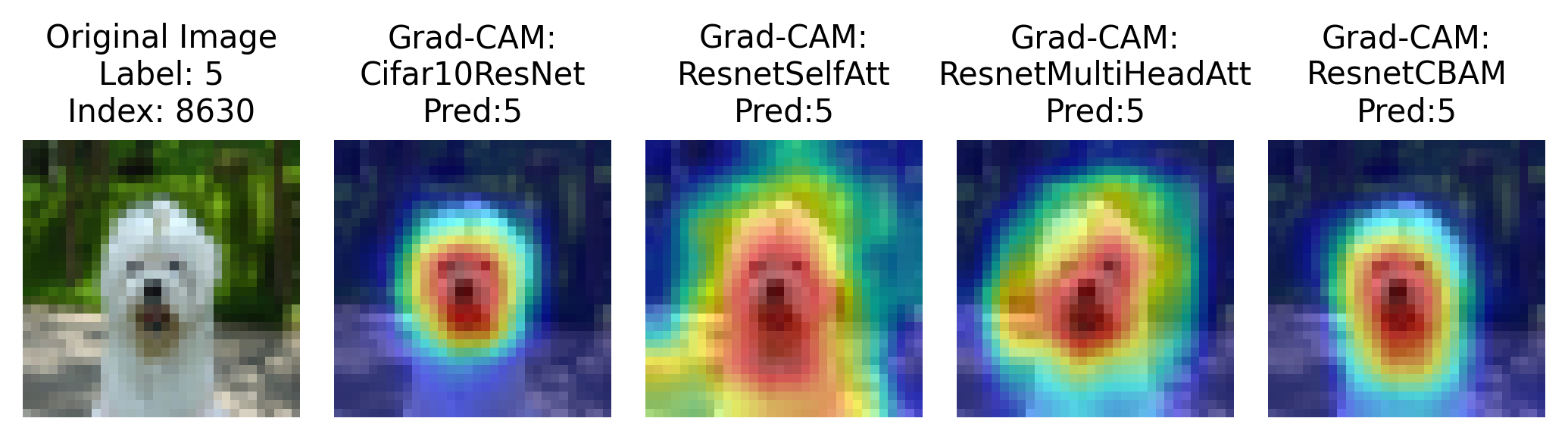}
\includegraphics[width=0.8\textwidth]{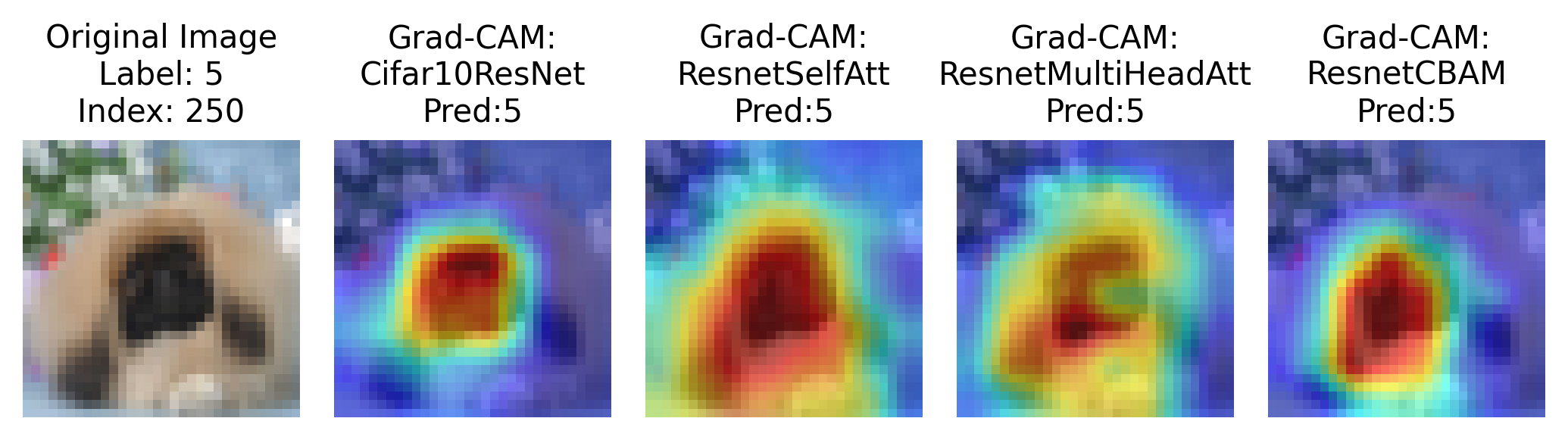}
\includegraphics[width=0.8\textwidth]{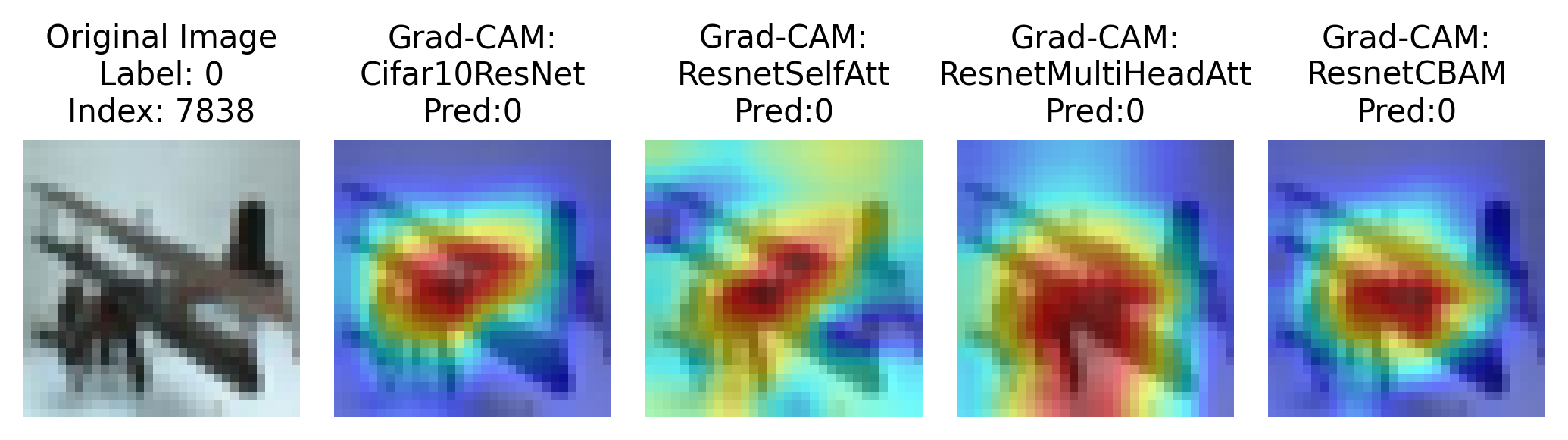}
\includegraphics[width=0.8\textwidth]{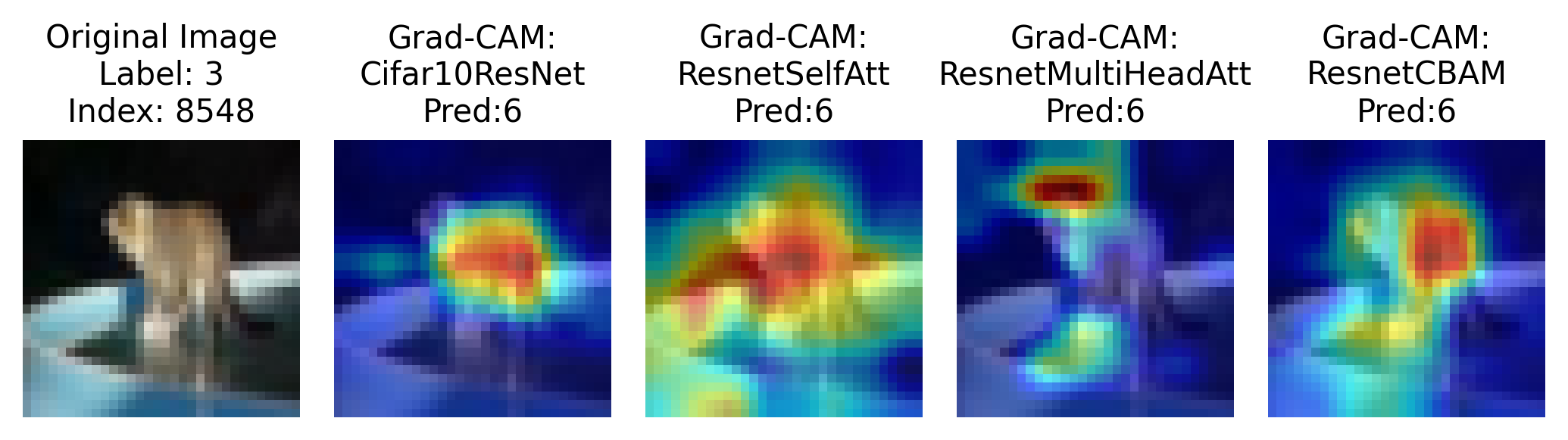}
\caption{More GradCAM results on CIFAR-10 images.}
\label{fig:gradcam-cifar-appen}
\end{figure}

\clearpage

For MNIST, we can see that the baseline model in all cases extracts relevant features. In a similar vein, the SelfAtt model tries to capture the global context whereas MHA's distributed representation tries to do it more effectively. Followed by the CBAM block that tends to enhance existing baseline results with more sharper focus.

\begin{figure}[h]
\centering
\includegraphics[width=0.8\textwidth]{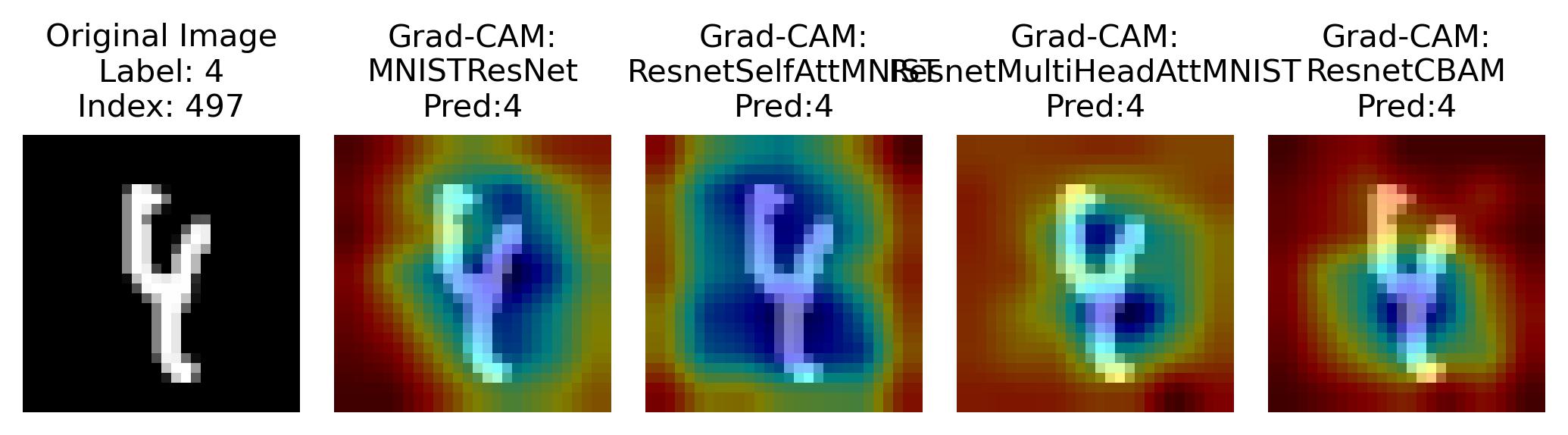}
\includegraphics[width=0.8\textwidth]{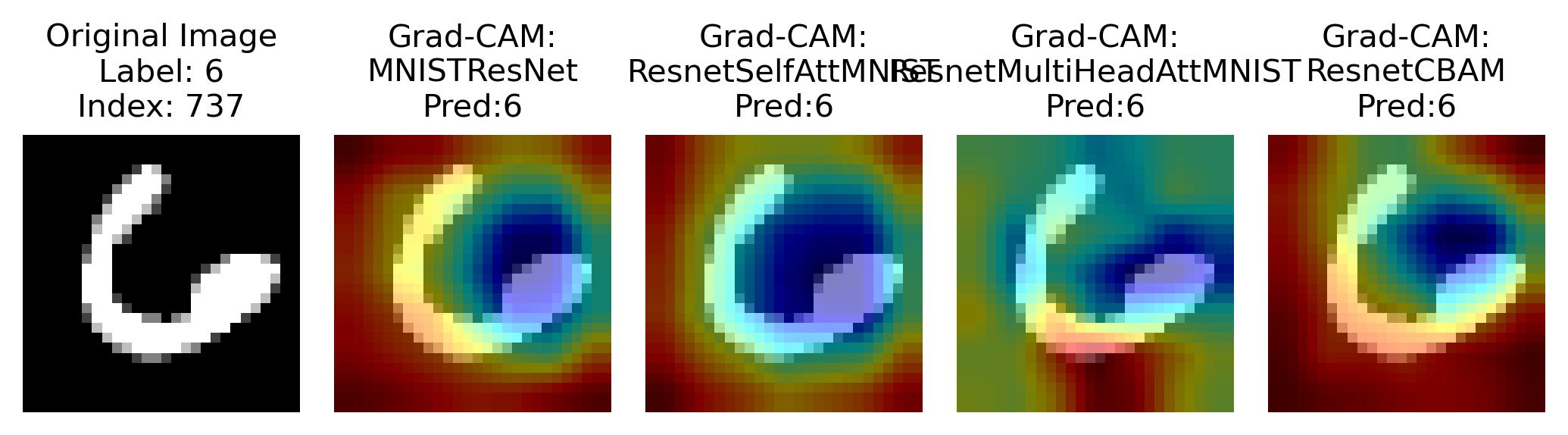}
\includegraphics[width=0.8\textwidth]{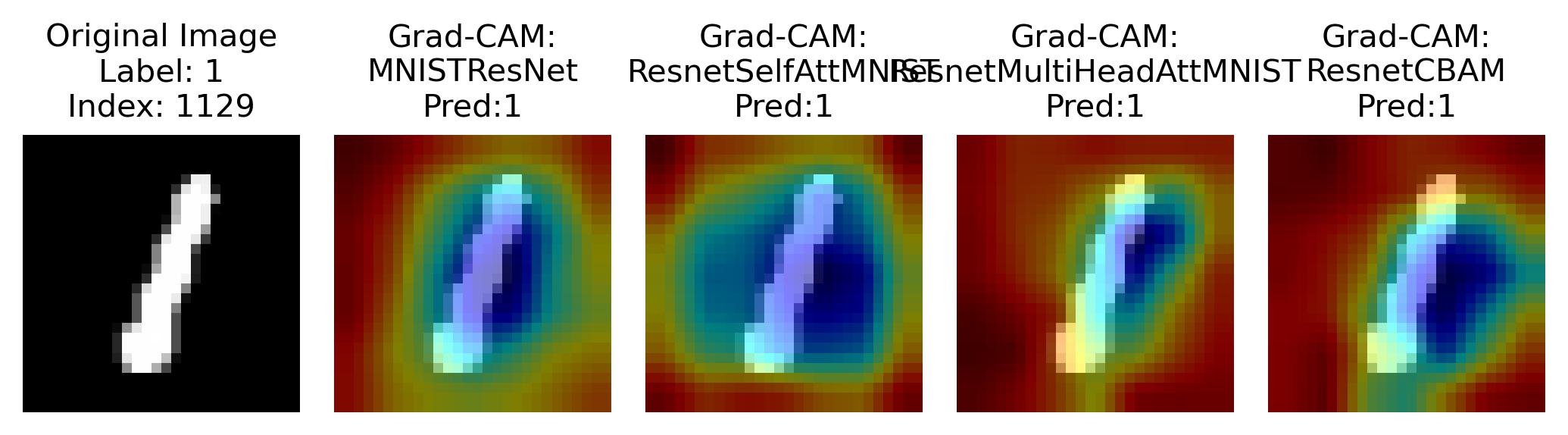}
\includegraphics[width=0.8\textwidth]{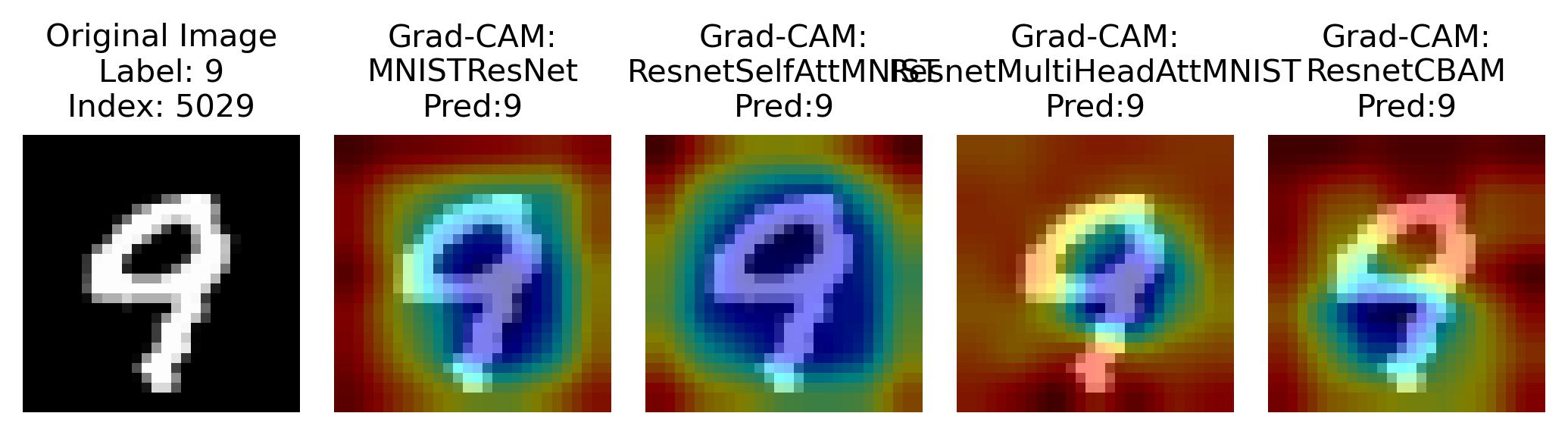}
\caption{More GradCAM results on MNIST images.}
\label{fig:gradcam-mnist-appen}
\end{figure}

%% file: 9-work-division.tex
\clearpage
% \FloatBarrier
\onecolumn
\section{Work Division}

Please add a section on the delegation of work among team members at the end of the report, in the form of a table and paragraph description. This and references do \textbf{NOT} count towards your page limit. An example has been provided in Table \ref{tab:contributions}.

\begin{table*}[h]
\begin{center}
\begin{tabular}{|l|c|p{6cm}|}
\hline
Student Name & Contributed Aspects & Details \\
\hline
Julian Glattki & Machine Learning Pipeline and ResNet reimplementation & Setup pipeline for entire Machine Learning flow using PyTorch, Skorch and MLFlow. Rebuilt the original ResNet20 from \cite{deepresiduallearningimage} on CIFAR-10.
% \\
% \newline & \newline & \newline
\\
Nikhil Kapila & Self and MultiHead Self Attention. Created utilities. & Implementation and experimentation of self and multi-head attention mechanisms (and 1 unweighted) on CIFAR10 and MNIST, UML diagrams, Overall model architecture diagrams, made utilities to load MLFlow models, plot graphs and make GradCAM inferences. ResNet baseline on MNIST. Bug fixes in Pipeline. Discussed \& reported experimentation findings.
\\
% \newline & \newline & \newline
% \\
Tejas Rathi & Convolution Block Attention Module (CBAM) implementation & Implemented the CBAM module, referring to the methodology outlined in the paper \cite{CBAM}, while introducing modifications to its placement within the ResNet20 (our baseline). Experimented with three different reduction ratios to study effectiveness of attention. Tuned CBAM based models with different hyperparameters. Discussed \& reported experimentation findings.\\
\hline
\end{tabular}
\end{center}
\caption{Contributions of team members.}
\label{tab:contributions}
\end{table*}